# DGT-TM: A freely Available Translation Memory in 22 Languages

Ralf Steinberger♪, Andreas Eisele♫, Szymon Klocek♫, Spyridon Pilos♫ & Patrick Schlüter♫

(♪) European Commission
Joint Research Centre (JRC)
Ispra (VA), Italy
Ralf.Steinberger@jrc.ec.europa.eu

(♫) European Commission
Directorate General for Translation (DGT)
Luxembourg, Luxembourg
Firstname.Lastname@ec.europa.eu

**Abstract**

The European Commission's (EC) *Directorate General for Translation*, together with the EC's *Joint Research Centre*, is making available a large translation memory (TM; i.e. sentences and their professionally produced translations) covering twenty-two official European Union (EU) languages and their 231 language pairs. Such a resource is typically used by translation professionals in combination with TM software to improve speed and consistency of their translations. However, this resource has also many uses for translation studies and for language technology applications, including Statistical Machine Translation (SMT), terminology extraction, Named Entity Recognition (NER), multilingual classification and clustering, and many more. In this reference paper for DGT-TM, we introduce this new resource, provide statistics regarding its size, and explain how it was produced and how to use it.

**Keywords:** Translation Memory; 231 language pairs; European Commission.

## 1. Introduction and Motivation

In recent years, the European Commission has released a number of large-scale multilingual linguistic resources. These are the parallel corpus *JRC-Acquis* (Steinberger et al. 2006), the translation memory *DGT-TM* (2007), the named entity recognition and normalisation resource *JRC-Names* (Steinberger et al. 2011) and the *JRC Eurovoc Indexer* JEX (Steinberger et al. 2012). The European Commission is now releasing an update to *DGT-TM* that contains documents published between the years 2004 and 2010 and that is two times larger than the release from 2007. From now on, it is planned to release DGT-TM data yearly. This release, covering data produced until the end of the year 2010, is called *DGT-TM Release 2011*.

This effort to make multilingual data available and to thus support the development of Language Technology solutions is related to the affirmation that multilinguality is one of the basic principles of the EU, guaranteeing cultural and linguistic diversity. By giving the EU citizen access to legislative and policy proposals in all the official EU languages, translation and cross-lingual information access technology contribute to making the EU more transparent, egalitarian, accountable and democratic.

The release can furthermore be seen in the context of Directive 2003/98/EC of the European Parliament and of the Council on the re-use of public sector information.[1] This directive recognises that public sector information such as multilingual collections of documents can be an important primary material for digital content products and services, that their release may have an impact on cross-border exploitation of information, and that it may thus have a positive effect on an unhindered competition in the EU's internal market. Related developments are that the European Institutions made publicly accessible the full-text database of EU law EUR-Lex[2], the database of EU terminology IATE[3], the multilingual wide-coverage thesaurus EuroVoc[4], plus other resources for translators[5] (EC&DGT 2008). EUR-Lex provides free access to European Union law and other documents considered to be public, written in all 23 official EU languages. The IATE website (*Inter-Active Terminology for Europe*) gives access to a database of EU inter-institutional terminology. IATE has been used in the EU institutions and agencies since 2004 for the collection, dissemination and shared management of EU-specific terminology. EuroVoc is a multilingual thesaurus originally built specifically for the manual indexing and retrieval of multilingual documentary information of the EU institutions, but it is now much more widely used, e.g. by the libraries of many national governments in the EU. It is a multi-disciplinary thesaurus covering fields that are sufficiently wide-ranging to encompass both Community and national points of view, with a certain emphasis on parliamentary activities. EuroVoc is a controlled set of vocabulary which can also be used outside the EU institutions, particularly by parliaments. JRC has publicly released its *JRC EuroVoc indexer* software JEX (Steinberger et al. 2012), which multi-label classifies documents according to the multilingual EuroVoc thesaurus and thus allows establishing links between documents written in different languages.

While the original textual data contained in DGT-TM, produced by the European Institutions and the EU Member States, consists of texts and their translation written for mostly legal purposes, the text collection has many practical uses outside the legal domain (see Section 2). DGT's *Information Technology Unit* and its *Language Applications Sector*, who are responsible for the development and maintenance of translation aids such as machine translation, translation memories, and reference and documentation search facilities (EC&DGT 2008), decided to make use of the EU's text collections to enhance their TMs. Now that the original full-text data has been sentence-aligned and added to DGT's TM, the resource is being released publicly.

---

[1] For details and to read the full text of the regulation, see http://eur-lex.europa.eu/LexUriServ/LexUriServ.do?uri=CELEX:32003L0098:EN:NOT. All URLs were last visited in October 2011.

[2] See http://eur-lex.europa.eu/.
[3] See http://iate.europa.eu/.
[4] See http://eurovoc.europa.eu/.
[5] See http://ec.europa.eu/dgs/translation/publications/.

We will now list possible Language Technology uses of multilingual TMs and similar resources (Section 2), describe DGT-TM and how it was produced (3), and give some practical details on its usage (4).

## 2. Possible uses of DGT-TM

The value of collections of parallel texts or sentences is widely accepted and there have been efforts to make such collections available publicly (e.g. Macklovitch et al., 2000; Koehn, 2005), although some of them are only accessible via web interfaces (e.g. Parasol, 2011). The number of highly multilingual parallel collections is relatively low, with *EuroParl* (Koehn 2005; 21 languages in the 2011 update), DGT-TM (2007) and JRC-Acquis (Steinberger et al., 2006) (22 languages each) being among the most highly multilingual ones. There have also been efforts to make various third-party resources available via a single website (Tiedemann, 2009).

One area that crucially depends on parallel data is the creation of models for Statistical Machine Translation (SMT). The initial work on SMT made use of proceedings of Canadian parliament debates (Hansard) available in English and French. Since 2001, SMT work funded by DARPA focused on translation from Chinese and Arabic into English, for which models were trained using large parallel corpora, such as United Nations publications. The EuroParl corpus with its originally 11 languages allowed the creation of SMT systems for up to 110 language pairs (Koehn 2005) and provided a crucial pre-condition for work in projects like *EuroMatrix* and *EuroMatrix Plus*.[6] The publication of the JRC-Acquis in 2006 enabled the creation of SMT systems for 462 European language pairs (Koehn et al. 2009).

The value of not only multi-monolingual, but *parallel* resources (corpora, dictionaries, tools) cannot be estimated high enough because it makes the effort to develop, train and test multilingual text mining tools more efficient and comparable (Steinberger, 2011). Parallel collections of sentences have been used, among other things, for the following tasks:

- Producing multilingual lexical and semantic resources such as dictionaries and ontologies;
- Training and testing information extraction software;
- Annotation projection across languages for Named Entity Recognition (Ehrmann et al. 2011), sentiment analysis (Josef Steinberger et al., 2011), multi-document summarisation (Turchi et al. 2010), part-of-speech annotation, word sense disambiguation, and more (Yarowsky et al. 2001); Annotation projection allows saving annotation time and it creates more comparable resources for many languages;
- Cross-lingual plagiarism detection (Potthast et al. 2011)
- Multilingual and cross-lingual clustering and classification (Wei et al. 2008).
- More generally, creation of multilingual semantic space in Lexical Semantic Analysis (LSA; Landauer & Littman 1991), Kernel Canonical Correlation Analysis (KCCA; Vinokourov et al. 2002), etc.

When the full text (i.e. information on the ordering of the sentences in the document) is available, further uses are possible:
- Translation studies, annotation projection for co-reference resolution, discourse analysis, comparative language studies;
- Checking translation consistency automatically;
- Making use of full-text information to improve SMT;
- Testing and benchmarking alignment software (for sentences, words, etc.);

## 3. Details on DGT-TM-2011

In 2011, DGT decided to import a large number of official EU documents with the purpose of adding them to the TM used by DGT's translators. For that purpose, the existing full-text document collections were automatically sentence-aligned. The next sub-sections specify the text type of the documents, discuss the translation quality of the corpus, describe the alignment procedure and provide statistics on the resulting multi-language-aligned corpus.

### 3.1 Documents contained in DGT-TM-2011

Since January 1968 the EC's *Official Journal* (OJ) has been published in two separate series: L (Legislation) and C (Information and Notices). The documents selected for inclusion in DGT-TM-2011 are those that are part of the L-Series, published between 2004 and 2010, inclusive. The documents are thus part of the *Acquis Communautaire* (i.e. the common body of EU law). Documents that were already contained in the DGT-TM release of the year 2007 were excluded, so as to avoid data overlap. New data releases are planned annually. DGT also plans to process the C-series documents in the future.

### 3.2 Quality of the translations

DGT-TM contains official legal acts. Starting from the drafting language (in 2008, 72% of all documents were drafted in English and 11% in French), the versions in the other 22 languages are produced by highly qualified human translators specialised in specific subject domains. All documents are linguistically and legally checked during a multi-step revision process. The quality is controlled at the level of each translation service, by legal services and by the EC's Publications Office (PO). There is much focus on ensuring terminology consistency, which includes involvement of the public administrations of the EU Member States. EU translators work in a cutting-edge IT environment, with many custom-built enhancements aimed at streamlining the work and ensuring quality and consistency. While it is in the nature of translation that its quality is always arguable, it can be assumed that the translation quality in DGT-TM on average is of a very high standard.

---

[6] See http://www.euromatrix.net/ and http://www.euromatrixplus.net/.

| Lang. | No. of TUs DGT-TM-2007 | No. of TUs DGT-TM-2011 | No. of Words | Words/TU | No. of ; and : | No. of chars | Chars/TU | Std.dev. Chars/TU | Size on disk, GiB (with En) |
|---|---|---|---|---|---|---|---|---|---|
| | 2007 | 2011 | | | | | | | |
| BG | 708,658 | 454,812 | 8,071,010 | 17.75 | 84,694 | 52,628,776 | 116 | 116 | 325,793,830 |
| CS | 890,025 | 1,985,152 | 28,612,679 | 14.41 | 330,484 | 187,049,247 | 94 | 97 | 1,316,825,816 |
| DA | 433,871 | 1,997,649 | 29,970,762 | 15.00 | 262,054 | 203,699,794 | 102 | 105 | 1,360,698,542 |
| DE | 532,668 | 1,922,568 | 29,917,127 | 15.56 | 320,131 | 218,675,436 | 114 | 116 | 1,361,935,620 |
| EL | 371,039 | 1,901,490 | 33,254,233 | 17.49 | 235,087 | 220,773,430 | 116 | 119 | 1,355,099,308 |
| EN | 2,187,504 | 2,286,514 | 38,967,261 | 17.04 | 409,442 | 236,139,765 | 103 | 112 | n/a |
| ES | 509,054 | 1,907,649 | 36,762,677 | 19.27 | 342,712 | 223,057,864 | 117 | 120 | 1,363,854,954 |
| ET | 1,047,503 | 1,867,786 | 22,668,156 | 12.14 | 326,140 | 179,007,570 | 96 | 98 | 1,253,164,440 |
| FI | 514,868 | 1,881,558 | 23,703,399 | 12.60 | 561,036 | 202,462,070 | 108 | 109 | 1,313,289,534 |
| FR | 1,106,442 | 1,853,773 | 37,121,869 | 20.03 | 343,383 | 221,166,864 | 119 | 119 | 1,345,198,486 |
| HU | 1,159,975 | 1,869,246 | 26,947,447 | 14.42 | 358,951 | 204,113,757 | 109 | 112 | 1,307,689,922 |
| IT | 542,873 | 1,926,532 | 34,335,360 | 17.82 | 357,981 | 219,557,996 | 114 | 116 | 1,364,689,856 |
| LT | 1,126,255 | 1,867,176 | 25,329,246 | 13.57 | 327,025 | 188,246,311 | 101 | 102 | 1,272,496,876 |
| LV | 1,120,835 | 1,859,781 | 25,484,395 | 13.70 | 302,593 | 183,064,221 | 98 | 99 | 1,260,388,126 |
| MT | 1,021,855 | 461,865 | 8,793,006 | 19.04 | 84,811 | 54,472,604 | 118 | 117 | 332,512,102 |
| NL | 502,557 | 1,914,628 | 32,970,482 | 17.22 | 363,964 | 218,286,026 | 114 | 117 | 1,356,588,864 |
| PL | 1,052,136 | 1,879,469 | 28,568,028 | 15.20 | 353,199 | 204,854,514 | 109 | 111 | 1,310,709,496 |
| PT | 945,203 | 1,922,585 | 35,559,701 | 18.50 | 343,195 | 214,970,957 | 112 | 114 | 1,351,988,416 |
| RO | 650,735 | 470,303 | 8,379,837 | 17.82 | 92,590 | 55,175,428 | 117 | 119 | 336,809,388 |
| SK | 1,065,399 | 1,894,676 | 28,064,088 | 14.81 | 320,621 | 187,085,316 | 99 | 101 | 1,280,611,526 |
| SL | 1,026,668 | 1,903,453 | 27,902,085 | 14.66 | 336,085 | 179,652,798 | 94 | 96 | 1,269,297,360 |
| SV | 555,362 | 1,934,964 | 29,236,484 | 15.11 | 218,153 | 198,223,221 | 102 | 104 | 1,323,916,224 |
| All | 19,071,485 | 37,963,629 | 600,619,332 | | 6,674,331 | 4,052,363,965 | | | |

**Table 1.** Size of the data in DGT-TM Release 2011, covering the years 2004 until 2010. For comparison, the first column shows the number of TUs of the 2007 release. Size is expressed in number of TUs, words and characters (chars). The number of colons and semi-colons gives an approximate indication on how many TUs are sentence parts. The size on disk, expressed in Gibibytes, refers to the bilingual TMX files with English as the second language.

### 3.3 Alignment of translation units

The manually translated full-text documents were thus the input to DGT's processing, which resulted in a TM with a collection of sentences and their translations. Rather than speaking of *sentences* as the basic alignment unit, it is more useful to speak of *Translation Units* (TUs). Translation Units are mostly traditional sentences, but they also contain titles and section headings, and they may also be sentence parts separated by colons or semi-colons. The usage of semi-colons to separate even longer, multi-paragraph sections is very common in legal jargon (between 16% and 17% of all TUs; see also **Table 1**). However, for all intended usages of this resource, the TUs in DGT-TM can be considered equivalent to sentences. We thus use both terms synonymously.

TUs in the parallel corpus were aligned between English and each of the other 21 languages. Alignments between language pairs not involving English are thus indirect and are produced by the accompanying software tool, exploiting the alignments of both languages with English.

DGT's in-house alignment tool is tuned to dealing with the specific features of EC and EU documents. The alignment algorithm makes use of anchors such as numbers and text numbering, as well as images and other non-linguistic information. Text numbering is used to first define zones, within which sentences can be aligned. The other clues are then used to strengthen the alignments. In the absence of any external clues, the system uses character count statistics on the typical relative length of sentences across languages.

Exact numbers for the alignment quality evaluation are not available, but the alignment was tested in a production setting where translators were confronted with the automatically aligned translations. They were encouraged to notify any alignment errors, and where such errors were found, they were used to improve the alignment algorithm. Altogether, the alignment quality was found to be very good, with only few errors.

### 3.4 Statistics about DGT-TM Version 2011

The result of the segmentation of all available documents in each language is a collection of about 38 million TUs in the 22 official EU languages (see **Table 1**). There are about 1.9 million TUs per language. Bulgaria and Romania joined the EU only in 2007 and Maltese translations were not obligatory in the first three years of Malta's EU membership (2004-2007), which explains the smaller number of TUs for these three languages. Irish Gaeilge (GA) became the 23rd official EU language in 2007, but the EU Institutions are currently exempt from the obligation to draft all acts in Irish.[7] The number of Irish documents was too small to be included in this version of DGT-TM.

**Table 2** shows the number of parallel TUs for each of the 231 language pairs.

Note that, while the DGT-TM releases of the years 2007 and 2011 do not have any full documents in common, they do share many individual sentences. An analysis of the English-Danish sets of sentences in both collections revealed that just over 3.5% of the combined sentences are exact duplicates, which justifies the usage of TMs. However, TMs can even exploit partially matching sentences so that translators can definitely benefit from TMs of the size of DGT-TM. There are also duplicate TUs within this new release. These are simply sentences or headers that occur repeatedly, across documents. While the majority of these repeated TUs have been excluded when creating this TM, others (especially from the earlier years) are included: There are 2275 TUs that are identical across all 22 languages, 13345 that are identical for ten or more languages, and 330,158 TUs that are identical for any language pair.

## 4. Downloading and using DGT-TM

The resource and the software tool can be downloaded from http://langtech.jrc.ec.europa.eu/JRC_Resources.html. The format of the DGT-TM-2011 files and the accompanying software are identical to those of the DGT-TM release in 2007.

### 4.1 Downloading DGT-TM

The corpus has been split into 25 individual zip file packages with up to 100MB each. Each zip file contains many TMX-files identified by the EUR-Lex identifier of the underlying documents and a file list in plain text format specifying the languages in which the documents are available. It is not necessary to unzip the files as the accompanying extraction program will access the data in the zip files directly.

The texts for the different languages are spread over the various zip files so that users will need to download all files even if they only want all the parallel sentences for a single language pair. Downloading only a subset of the zip files is possible, but it will result in producing only a subset of the parallel corpus.

The characters are encoded in UTF-16 Little Endian and the documents are in TMX format, which is a widely used TM format. For backwards compatibility, the header mentions TMX format 1.1, but the files are also compliant with TMX 1.4b.

### 4.2 Extracting parallel sentences from DGT-TM

In order to produce parallel sentence collections for any of the 231 possible language pairs involving the 22 input languages, users need to use the included extraction program *TMXtract* (with the extension *.exe* or *.jar*, depending on the operating system) and copy it into the same directory as the zip files with the data. This software has not changed since DGT-TM (2007). The program is distributed in two versions: a version with graphical user interface for the Windows operating system, consisting of two files: the program file and the library, and a machine-independent command line version in java byte code that can be run on any machine supporting a Java runtime of version 1.4 or newer. Users can produce bilingual extractions for any language pair by following the explicit instructions available on the download site.

Note that the sequence in the extracted files is not necessarily the same as in the underlying documents, and redundancies of text segments like "*Article 1*" are inevitable. The documents in the files are identified by the document number (Numdoc) of the original legislative document in the Eur-Lex database, but it should be noted that these documents have been modified during the pre-processing steps.

### 4.3 Conditions of use

The DGT-TM database is the exclusive property of the European Commission. DGT-TM can be re-used and disseminated, free of charge, without the need to make an individual application, and both for commercial and non-commercial purposes, within the limits set by the Commission Decision of 12 December 2011 on the re-use of Commission documents ("Re-use Decision"), published in the Official Journal of the European Union L330 of 14 December 2011, pages 39 to 42.

Any natural or legal person who re-uses the DGT-TM documents, in accordance with the conditions laid down in the Re-use Decision, is obliged to state the source of the documents used: the website address, the date of the latest update and the fact that the European Commission retains ownership of the data.

The software distributed with DGT-TM, necessary for its exploitation/extraction, must be used in accordance with the conditions laid down in the licence GPL 2.0.

The database and the accompanying software are made available without any guarantee. The Commission is not liable for any consequence stemming from the re-use. Moreover, the Commission is not liable for the quality of the alignment nor the correctness of the data provided.

---

[7] http://publications.europa.eu/code/pdf/370000en.htm#fn4-2

For more detailed information please refer to the section on the conditions of use of the DGT-TM website.

## 5. Summary and future work

Following the release of the JRC-Acquis (Steinberger et al. 2006) and the first version of the DGT-TM in 2007, DGT and JRC have released, in 2012, a new and larger collection of sentences and their translations in up to 22 different languages. This latest release, named *DGT-TM Version 2011*, includes the documents from the L-Series (Legislation) of the EU's *Official Journal* published between 2004 and 2010. It is planned to make future releases of the TM annually.

TMs consist of individual sentences and sentence-like fragments that do not allow reproducing the original text while, for some purposes, it would be beneficial to have access to the full texts and thus to the order of sentences in the documents. It is therefore planned to also release a full-text version of the parallel documents in the same set of languages.

|    | BG      | CS        | DA        | DE        | EL        | EN        | ES        | ET        | FI        | FR        | HU        | IT        | LT        | LV        | MT      | NL        | PL        | PT        | RO      | SK        | SL        | SV      |
|----|---------|-----------|-----------|-----------|-----------|-----------|-----------|-----------|-----------|-----------|-----------|-----------|-----------|-----------|---------|-----------|-----------|-----------|---------|-----------|-----------|---------|
| BG | 0       | 439,097   | 435,715   | 437,548   | 435,884   | 454,812   | 435,857   | 440,432   | 439,424   | 425,399   | 439,081   | 433,990   | 442,903   | 441,490   | 437,442 | 437,445   | 440,216   | 437,009   | 437,848 | 441,007   | 438,657   | 437,931 |
| CS | 439,097 | 0         | 1,915,711 | 1,829,231 | 1,794,764 | 1,985,152 | 1,814,280 | 1,811,097 | 1,782,057 | 1,766,803 | 1,806,608 | 1,836,453 | 1,814,318 | 1,784,306 | 450,562 | 1,834,845 | 1,822,668 | 1,835,650 | 458,062 | 1,860,098 | 1,861,483 | 1,846,427 |
| DA | 435,715 | 1,915,711 | 0         | 1,853,912 | 1,815,705 | 1,997,649 | 1,838,672 | 1,795,201 | 1,802,003 | 1,790,403 | 1,786,668 | 1,858,951 | 1,795,222 | 1,766,224 | 447,641 | 1,862,888 | 1,812,145 | 1,858,114 | 455,394 | 1,832,874 | 1,836,581 | 1,879,481 |
| DE | 437,548 | 1,829,231 | 1,853,912 | 0         | 1,808,874 | 1,922,568 | 1,813,671 | 1,769,227 | 1,817,949 | 1,756,161 | 1,799,292 | 1,810,200 | 1,766,945 | 1,782,289 | 448,896 | 1,816,324 | 1,771,751 | 1,808,538 | 456,191 | 1,789,165 | 1,788,473 | 1,820,903 |
| EL | 435,884 | 1,794,764 | 1,815,705 | 1,808,874 | 0         | 1,901,490 | 1,817,465 | 1,788,642 | 1,819,777 | 1,762,445 | 1,783,066 | 1,815,372 | 1,788,422 | 1,791,353 | 436,180 | 1,813,608 | 1,775,275 | 1,813,931 | 437,866 | 1,788,118 | 1,790,530 | 1,816,039 |
| EN | 454,812 | 1,985,152 | 1,997,649 | 1,922,568 | 1,901,490 | 0         | 1,907,649 | 1,867,786 | 1,881,558 | 1,853,773 | 1,869,246 | 1,926,532 | 1,867,176 | 1,859,781 | 461,865 | 1,914,628 | 1,879,469 | 1,922,585 | 470,303 | 1,894,676 | 1,903,453 | 1,934,964 |
| ES | 435,857 | 1,814,280 | 1,838,672 | 1,813,671 | 1,817,465 | 1,907,649 | 0         | 1,791,678 | 1,799,913 | 1,786,922 | 1,784,610 | 1,843,625 | 1,791,951 | 1,765,819 | 447,292 | 1,837,035 | 1,799,676 | 1,846,788 | 457,140 | 1,811,521 | 1,815,078 | 1,834,693 |
| ET | 440,432 | 1,811,097 | 1,795,201 | 1,769,227 | 1,788,642 | 1,867,786 | 1,791,678 | 0         | 1,779,175 | 1,740,694 | 1,777,466 | 1,791,953 | 1,812,810 | 1,795,401 | 439,883 | 1,801,510 | 1,797,042 | 1,794,747 | 441,354 | 1,813,146 | 1,813,647 | 1,803,097 |
| FI | 439,424 | 1,782,057 | 1,802,003 | 1,817,949 | 1,819,777 | 1,881,558 | 1,799,913 | 1,779,175 | 0         | 1,745,893 | 1,798,155 | 1,799,433 | 1,778,441 | 1,801,699 | 439,501 | 1,802,881 | 1,764,523 | 1,796,426 | 441,020 | 1,776,837 | 1,778,169 | 1,811,762 |
| FR | 425,399 | 1,766,803 | 1,790,403 | 1,756,161 | 1,762,445 | 1,853,773 | 1,786,922 | 1,740,694 | 1,745,893 | 0         | 1,730,303 | 1,791,484 | 1,742,548 | 1,714,554 | 437,748 | 1,788,758 | 1,745,611 | 1,791,452 | 444,441 | 1,758,807 | 1,762,433 | 1,789,877 |
| HU | 439,081 | 1,806,608 | 1,786,668 | 1,799,292 | 1,783,066 | 1,869,246 | 1,784,610 | 1,777,466 | 1,798,155 | 1,730,303 | 0         | 1,783,452 | 1,776,692 | 1,805,137 | 450,115 | 1,782,519 | 1,780,089 | 1,781,830 | 457,899 | 1,798,236 | 1,801,621 | 1,790,040 |
| IT | 433,990 | 1,836,453 | 1,858,951 | 1,810,200 | 1,815,372 | 1,926,532 | 1,843,625 | 1,791,953 | 1,799,433 | 1,791,484 | 1,783,452 | 0         | 1,790,132 | 1,763,509 | 445,685 | 1,838,476 | 1,799,310 | 1,843,212 | 454,398 | 1,813,116 | 1,816,100 | 1,839,632 |
| LT | 442,903 | 1,814,318 | 1,795,222 | 1,766,945 | 1,788,422 | 1,867,176 | 1,791,951 | 1,812,810 | 1,778,441 | 1,742,548 | 1,776,692 | 1,790,132 | 0         | 1,786,017 | 438,743 | 1,804,459 | 1,809,896 | 1,797,388 | 439,656 | 1,818,576 | 1,817,882 | 1,803,752 |
| LV | 441,490 | 1,784,306 | 1,766,224 | 1,782,289 | 1,791,353 | 1,859,781 | 1,765,819 | 1,795,401 | 1,801,699 | 1,714,554 | 1,805,137 | 1,763,509 | 1,786,017 | 0         | 441,354 | 1,769,469 | 1,766,902 | 1,766,452 | 442,070 | 1,782,117 | 1,784,057 | 1,773,567 |
| MT | 437,442 | 450,562   | 447,641   | 448,896   | 436,180   | 461,865   | 447,292   | 439,883   | 439,501   | 437,748   | 450,115   | 445,685   | 438,743   | 441,354   | 0       | 449,372   | 446,214   | 449,179   | 450,736 | 451,413   | 450,545   | 449,312 |
| NL | 437,445 | 1,834,845 | 1,862,888 | 1,816,324 | 1,813,608 | 1,914,628 | 1,837,035 | 1,801,510 | 1,802,881 | 1,788,758 | 1,782,519 | 1,838,476 | 1,804,459 | 1,769,469 | 449,372 | 0         | 1,805,855 | 1,857,164 | 454,995 | 1,837,352 | 1,835,102 | 1,864,568 |
| PL | 440,216 | 1,822,668 | 1,812,145 | 1,771,751 | 1,775,275 | 1,879,469 | 1,799,676 | 1,797,042 | 1,764,523 | 1,745,611 | 1,780,089 | 1,799,310 | 1,809,896 | 1,766,902 | 446,214 | 1,805,855 | 0         | 1,803,353 | 455,153 | 1,824,590 | 1,823,905 | 1,819,577 |
| PT | 437,009 | 1,835,650 | 1,858,114 | 1,808,538 | 1,813,931 | 1,922,585 | 1,846,788 | 1,794,747 | 1,796,426 | 1,791,452 | 1,781,830 | 1,843,212 | 1,797,388 | 1,766,452 | 449,179 | 1,857,164 | 1,803,353 | 0         | 457,703 | 1,834,652 | 1,838,650 | 1,857,767 |
| RO | 437,848 | 458,062   | 455,394   | 456,191   | 437,866   | 470,303   | 457,140   | 441,354   | 441,020   | 444,441   | 457,899   | 454,398   | 439,656   | 442,070   | 450,736 | 454,995   | 455,153   | 457,703   | 0       | 459,852   | 458,835   | 457,180 |
| SK | 441,007 | 1,860,098 | 1,832,874 | 1,789,165 | 1,788,118 | 1,894,676 | 1,811,521 | 1,813,146 | 1,776,837 | 1,758,807 | 1,798,236 | 1,813,116 | 1,818,576 | 1,782,117 | 451,413 | 1,837,352 | 1,824,590 | 1,834,652 | 459,852 | 0         | 1,860,330 | 1,840,350 |
| SL | 438,657 | 1,861,483 | 1,836,581 | 1,788,473 | 1,790,530 | 1,903,453 | 1,815,078 | 1,813,647 | 1,778,169 | 1,762,433 | 1,801,621 | 1,816,100 | 1,817,882 | 1,784,057 | 450,545 | 1,835,102 | 1,823,905 | 1,838,650 | 458,835 | 1,860,330 | 0         | 1,843,135 |
| SV | 437,931 | 1,846,427 | 1,879,481 | 1,820,903 | 1,816,039 | 1,934,964 | 1,834,693 | 1,803,097 | 1,811,762 | 1,789,877 | 1,790,040 | 1,839,632 | 1,803,752 | 1,773,567 | 449,312 | 1,864,568 | 1,819,577 | 1,857,767 | 457,180 | 1,840,350 | 1,843,135 | 0       |

**Table 2.** Number of TUs per language pair.

459